\documentclass[11pt]{article}

\usepackage[OT1]{fontenc}
\usepackage{amsmath,amssymb,amsthm}
\usepackage{natbib}
\usepackage[usenames]{color}
\usepackage{xcolor}
\usepackage{graphicx}
\usepackage{booktabs}
\usepackage{multirow}
\usepackage{makecell}
\usepackage{array}
\usepackage{enumitem}
\usepackage{mathrsfs}
\usepackage{comment}
\usepackage{fullpage}
\usepackage{smile}
\usepackage[colorlinks,
            linkcolor=red,
            anchorcolor=blue,
            citecolor=blue,
            urlcolor=blue
            ]{hyperref}
\usepackage[protrusion=true,
            expansion=true,
            final,
            babel
            ]{microtype}

\makeatletter
\let\noteoldmaketitle\maketitle
\newcommand{\storedabstract}{}
\newcommand{\storedkeywords}{}
\newcommand{\TITLE}[1]{\title{#1}}
\newcommand{\AUTHOR}[1]{#1}
\newcommand{\AFF}[1]{\\\begin{tabular}[t]{c}\normalsize #1\end{tabular}}
\newcommand{\ARTICLEAUTHORS}[1]{\author{#1}}
\long\def\ABSTRACT#1{\gdef\storedabstract{#1}}
\newcommand{\KEYWORDS}[1]{\gdef\storedkeywords{#1}}
\newcommand{\HISTORY}[1]{}
\renewcommand{\maketitle}{%
  \noteoldmaketitle
  \begin{abstract}
  \storedabstract
  \end{abstract}
  \noindent\textbf{Keywords:} \storedkeywords\par\bigskip
}
\makeatother

\newcommand{\RUNAUTHOR}[1]{}
\newcommand{\RUNTITLE}[1]{}
\newcommand{\MANUSCRIPTNO}[1]{}
\newcommand{\VOLUME}[1]{}
\newcommand{\NO}[1]{}
\newcommand{\MONTH}[1]{}
\newcommand{\YEAR}[1]{}
\newcommand{\FIRSTPAGE}[1]{}
\newcommand{\LASTPAGE}[1]{}
\newcommand{\SHORTYEAR}[1]{}
\newcommand{\ISSUE}[1]{}
\newcommand{\LONGFIRSTPAGE}[1]{}
\newcommand{\DOI}[1]{}

\begin{document}



\TITLE{Bellman–Taylor Score Decoding for Markov Decision Processes with State-Dependent Feasible Action Sets}

\ARTICLEAUTHORS{%
\AUTHOR{Yi Chen, Rushuai Yang, Qiang Chen, Dongyan (Lucy) Huo}
\AFF{Department of Industrial Engineering and Decision Analytics\\ Hong Kong University of Science and Technology}
} 

\ABSTRACT{%
Many Markov decision processes (MDPs) in operations research have feasible actions that are state dependent and defined implicitly by various operational constraints. These features make it difficult to use standard deep reinforcement learning (DRL) algorithms, whose action interfaces typically assume either a fixed finite action catalog or a simple Euclidean space. Motivated by a Taylor expansion of the optimal action-value function, we propose \emph{Bellman--Taylor score decoding}, a framework that moves policy learning to a Euclidean score space while enforcing feasibility through an action decoder. The induced latent-score MDP then can be optimized by standard DRL algorithms without differentiating through the decoder. We provide a performance guarantee showing that the optimality gap of this approach decomposes into a structural approximation error and an algorithmic learning error. Lastly, we apply this framework to a queueing network control problem, where the policy essentially learns a state-dependent index-based dispatching rule. Numerical experiments show near-optimal performance in small instances and considerable improvements over benchmarks in larger systems.
}%


\KEYWORDS{Markov Decision Process, Deep Reinforcement Learning, Queueing Network Control}
\HISTORY{}

\date{}  
\maketitle
\section{Introduction}
\label{sec:introduction}

In recent years, deep reinforcement learning (DRL) has emerged as one of the most successful algorithmic paradigms for sequential decision-making. By combining reinforcement learning algorithms with expressive neural-network function approximation, DRL has achieved remarkable empirical success in many AI domains such as game playing, robotic control, and autonomous racing \citep{mnih2015human,Rajeswaran-RSS-18,wurman2022outracing,tang2025deep}. These successes suggest that DRL can be a powerful tool for learning complex control policies from simulation or interaction data, especially when explicit dynamic-programming solutions are computationally infeasible.

Many sequential decision-making problems in operations research (OR) can naturally be formulated as Markov decision processes (MDPs). Examples include inventory control, queueing control, and resource allocation. Classical dynamic programming provides a principled framework for such problems, but exact solution methods quickly become intractable as the state and action spaces grow. Approximate dynamic programming (ADP) and simulation-based policy optimization have therefore long played an important role \citep{bertsekas1996neuro,powell2011approximate}. From a mathematical perspective, ADP and DRL are closely connected: both seek tractable approximations to Bellman evaluation and improvement steps, either by approximating value functions, action-value functions, or policies. This connection has motivated a growing body of work applying DRL algorithms to solve large-scale OR problems. \citep{delarue2020reinforcement,dai2022queueing,gijsbrechts2022can,harsha2025deep,xu2025reinforcement,chen2026primal}.

Despite this natural connection, applying off-the-shelf DRL algorithms to operational MDPs is rarely plug-and-play. A critical challenge is the action interface. Standard DRL implementations typically assume that actions can be enumerated, sampled, and optimized over through a simple and fixed representation. In finite-action problems, a neural network often outputs one logit or value for each action in a fixed catalog. In continuous-action problems, the action is usually represented as a fixed-dimensional Euclidean vector, and the actor outputs either this vector or the parameters of a distribution over such vectors. Many OR problems do not fit either template. Their feasible actions are often state dependent, high dimensional, and implicitly defined by capacity, compatibility, and integrality constraints.

The action-interface difficulty is a critical reason why successful DRL applications in operational MDPs often require substantial case-specific algorithmic engineering. Researchers may need to design problem-specific action decompositions, feasibility corrections, masking rules, or specialized network architectures. These techniques can be effective, but they also reduce the extent to which standard DRL solvers can be used as reusable tools across operational MDPs. In this sense, the challenge is not only that OR problems are large or stochastic; it is also essential the natural action spaces of OR models often do not align with the action representations expected by standard DRL implementations.

The existing literature have made important progress toward addressing this issue. For example, action-representation methods embed large finite action sets into lower-dimensional spaces, allowing feedback from one action to generalize to similar actions \citep{chandak2019learning,dulacarnold2015deep}. Feasibility-preserving methods such as masking, action elimination, can prevent infeasible actions from being executed when a fixed action catalog or a simple feasible region is available \citep{huang2022closer,zahavy2018learn,pham2018optlayer}. A more recent line of work combines reinforcement learning with optimization-based action 
selection, either by embedding learned value information into an action selection problem or by incorporating an optimization layer inside the policy representation \citep{delarue2020reinforcement,harsha2025deep,xu2025reinforcement,hoppe2025structured}.

These approaches are valuable, but do not fully address the interface problem that arises when 
one wants to apply standard off-the-shelf DRL algorithms to MDPs with state-dependent feasible action sets. Action-representation methods are primarily designed for fixed finite action catalogs, whereas many feasible sets are defined implicitly by constraints and may be too large to enumerate. Masking and elimination also rely on an explicit catalog or a finite superset of actions. Optimization-layer methods typically treat the optimization score as a generic learned utility coefficient and require specialized training for the layer. These approaches either depend on a particular action representation, a 
problem-specific feasibility mechanism, or a customized optimization-based learning architecture. 

In this paper, we propose a novel action interface for MDPs with state-dependent feasible action sets, which we call the \emph{Bellman-Taylor score decoding}. The central idea is to standardize the learning interface rather than the operational action space. Instead of learning a policy directly over the irregular feasible action set, the policy learns a score vector in a Euclidean space. An action-decoder then maps this score into an implementable natural action by solving an optimization problem over the original feasible set. The decoder is motivated by a Taylor approximation of the post-action Bellman continuation value (or the optimal action-value function), where the learned score is intended to represent the marginal effect of the post-action system configuration on future value. 

This construction separates learning from feasibility. The DRL algorithm operates on a regular Euclidean score space, while feasibility, integrality, and combinatorial coupling are handled by the action decoder. Once the decoder is specified, the original MDP induces a latent-score MDP in which the action is the score vector. Standard continuous-action DRL algorithms, such as proximal policy optimization, can then directly be applied to this induced problem. Importantly, the decoder is solved only in the forward pass to convert a sampled score into a feasible natural action. The policy-gradient update does not require differentiating through the decoder. This is distinct with differentiable optimization-layer methods, where the optimizer is
part of the trainable computation graph and must be differentiated through or approximated
by a surrogate gradient.  

The proposed framework does not eliminate the inherent difficulty of operational MDPs. Instead, it confines model structure to the decoder, allowing the learning algorithm to remain standard. The model specifies the feasible action set, immediate reward, and a post-action representation of the system. The policy learns state-dependent scores, which the decoder translates into feasible operational actions. This creates a bridge between off-the-shelf continuous-action DRL solvers and MDPs with state-dependent, constrained, and often combinatorial actions. The trade-off is that the decoder-induced policy class is restricted: not every feasible policy in the original MDP can be represented by a score-decoding policy. Our theoretical analysis quantifies this trade-off. 

As an application of this framework, we investigate a queueing network control problem. Dynamic scheduling in multi-class, multi-pool queueing systems is a classical and practically important problem, with applications such as call centers and inpatient overflow management \citep{dai2019inpatient,chen2020survey}. Recent studies have also explored the use of DRL for such problems, but successful implementations often rely on problem-specific action decompositions, customized policy architectures, or variance-reduction techniques tailored to the queueing model. In contrast, our framework allows us to apply a standard PPO solver directly without introducing any additional problem-specific algorithmic engineering such as variance reduction, which are common and essential in peer works \citep{dai2022queueing,dong2025multiclass}. In this setting, the Bellman-Taylor score decoding essentially becomes a learned index-based dispatching rule: the RL policy learns state-dependent indices, and the action decoder selects the feasible dispatching action with the largest total score. Through simulation experiments, we show that the resulting decoded PPO policy outperforms various benchmarks across a range of queueing instances. 

In summary, our contributions are threefold.

\begin{itemize}
    \item  We introduce the Bellman-Taylor score-decoding algorithm framework for MDPs with state-dependent feasible action sets. The framework converts a constrained natural-action MDP into a latent-score MDP, so that standard continuous-action DRL algorithms can be applied without designing a problem-specific probability distribution over feasible actions. Feasibility is enforced exactly through the decoder.

    \item  We provide a structural performance guarantee showing that optimality gap of our approach can be decomposed into a structural approximation error and an algorithmic learning error from solving the induced latent-score MDP. The structural term is controlled by the local Taylor remainder of the post-action continuation value, clarifying when the restricted score-decoding policy class can approximate Bellman-greedy decisions well.

    \item  We apply the framework to a queueing network control problem. Our implementation uses a standard PPO solver on the latent-score MDP without queue-specific modifications to the learning algorithm or additional variance-reduction devices. The numerical study demonstrates the superior performance of our approach.
\end{itemize}

The remainder of the paper is organized as follows. Section~\ref{sec:related-literature} reviews related literature. Section~\ref{background} formulates the MDP setting and introduces background. Section~\ref{algorithm} introduces Bellman--Taylor score decoding framework and the PPO implementation, as well as performance guarantee. Section~\ref{inventory} investigates an inventory problem as a sanity check. Section~\ref{queue} presents the queueing network control case study. Section~\ref{conclusion} concludes lastly. 

\section{Related Literature}
\label{sec:related-literature}


\subsection{Deep Reinforcement Learning and Standard Action Interfaces}

For small-scale MDPs with finite state and action spaces, classical dynamic programming and exact MDP methods, including value iteration, policy iteration, and linear programming, provide exact solution approaches \citep{sutton1998reinforcement}. However, when the state and action spaces become large, these exact methods quickly become computationally prohibitive, due to the curse of dimensionality \citep{bellman1957dynamic,powell2011approximate}. This challenge has motivated a broad literature on approximation algorithms, with substantial progress in approximate dynamic programming \citep{bertsekas1996neuro,powell2011approximate}. Deep reinforcement learning can be viewed as a neural-network-based extension of this approximation paradigm, in which value functions or policies are represented by deep neural networks and learned from trajectory data. Owing to the strong approximation power of neural networks, DRL has achieved remarkable empirical success in a variety of benchmark problems in AI domains, such as game playing and robotic control \citep{mnih2015human,Rajeswaran-RSS-18,wurman2022outracing,tang2025deep}.

In general, classic DRL algorithms can be roughly divided into two categories: \emph{value-based} methods, including the deep $Q$-networks (DQN) and related variants \citep{mnih2015human}, which approximate the optimal action-value function by a neural network $Q_\theta(s,a)$ and select actions according to
$
a_t \in \arg\max_{a \in \mathcal{A}(s_t)} Q_\theta(s_t,a);
$
and \emph{policy-based} or \emph{actor-critic} methods, including the proximal policy optimization (PPO) and related variants \citep{schulman2017proximal}, which directly parameterize a policy as $\pi_\theta(a\mid s)$ and optimize the expected discounted return
$ J(\pi^\theta;\nu_0) $ over the parameter space of $\theta$ (which is often Euclidean) using stochastic gradient methods \citep{sutton1998reinforcement}. In both categories, deep neural networks serve as flexible function approximators and have demonstrated strong empirical performance in a variety of benchmark problems \citep{Rajeswaran-RSS-18,wurman2022outracing,tang2025deep}.

However, a common implicit assumption in standard DRL implementations is that they typically rely on a tractable action interface. In discrete-action settings, the action space is often a explicit finite action catalog of manageable size, and the neural network outputs one logit or $Q$-value for each candidate action \citep{mnih2013playing, mnih2015human}. In continuous-action settings, the action is often represented as a fixed-dimensional Euclidean vector, and the policy network outputs the parameters of a probability distribution, or directly outputs the action vector itself \citep{schulman2017proximal}. Thus, the neural-network architecture implicitly assumes that, once the current state is given, actions can be scored, sampled, or optimized over through a simple and fixed interface.


\subsection{Action Representations and Feasibility-Preserving Policies} \label{subsec:lit-action-representation}

When the feasible-action geometry becomes complex, or the feasible action sets vary across states, the standard feasibility mechanisms in DRL depend on how the action is represented. A related stream of work modifies the action parameterization so that large action spaces can be handled by standard function approximation. Action-embedding methods learn a low-dimensional representation of a large discrete action set and use this representation to generalize feedback across actions \citep{dulacarnold2015deep,chandak2019learning}. Other methods reduce the output dimension by exploiting factorized or hybrid action structure: action-branching architectures decompose high-dimensional discrete actions into componentwise decisions \citep{tavakoli2018action}, while parameterized-action and hybrid-action methods represent decisions through a fixed discrete action type together with continuous parameters \citep{hausknecht2016deep,li2021hyar}. These approaches are effective when the action space admits a fixed catalog, a fixed factorization, or a fixed hybrid template. They are less directly applicable when feasible actions are state dependent, implicitly specified by capacity and integrality constraints, and too large to enumerate.

Feasibility-preserving mechanisms in DRL are also closely related. Invalid-action masking removes infeasible actions from a fixed discrete catalog before sampling from the policy \citep{huang2022closer}. Action-elimination methods learn to discard invalid or undesirable actions, often using external elimination signals \citep{zahavy2018learn}. Projection and repair layers map an unconstrained action proposed by a neural network to a feasible action satisfying prescribed constraints \citep{pham2018optlayer}. Knowledge-compilation approaches encode valid combinatorial actions symbolically and learn policies supported only on the valid action set \citep{ling2023knowledge}. These methods are useful when feasibility can be represented through a tractable catalog, mask, projection, repair rule, or symbolic encoding.

Another approach handles combinatorial actions by factorizing a system-level decision into a sequence of smaller decisions. Atomic-action and autoregressive policies generate assignments one at a time, often using feasibility masks to rule out invalid local choices \citep{sun2024inpatient}. This idea avoids enumerating the full feasible action set and provides a scalable policy parameterization. It is natural for matching and routing problems, where a large assignment decision can be decomposed into individual customer-server matches. 

Bellman–Taylor score decoding provides an alternative solution strategy. The latent variable in our framework is not an embedding of an action, nor is it a componentwise action proposal to be repaired. It is a state-dependent marginal-value score motivated by a Taylor approximation of the action-value function. Given this score, the decoder selects an action by optimizing a Bellman-motivated surrogate over the original state-dependent feasible set. Thus, feasibility and action ranking are handled jointly: the decoder enforces the operational constraints exactly while using the learned score to rank feasible actions according to their estimated downstream value. This feature is convenient for MDPs whose feasible actions are naturally represented by optimization program or various constraints rather than by a fixed action catalog.

\subsection{Optimization-Oriented RL for Combinatorial Actions}
\label{subsec:lit-optimization-rl}

Our method is also connected to the area that combines RL with optimization-based action selection for combinatorial actions. In these methods, optimization is used not merely to repair infeasible actions, but to select actions from a structured feasible set. 

A family of approaches focuses on value-based RL algorithms, where a learned action-value function is embedded into a combinatorial action-selection problem at each decision step. For example, \citet{delarue2020reinforcement} formulate policy improvement as a mixed-integer optimization problem over the learned $Q$-function. \citet{harsha2025deep} combine neural value approximation with mathematical programming and sample-average approximation to solve large state-dependent inventory actions. \citet{xu2025reinforcement} embed an action-value network into a mixed-integer program for coupled restless bandits. These methods are highly expressive because the optimizer can directly exploit a nonlinear approximation of the downstream value function. Compared with our approach, these methods are computationally heavy, especially when the value function is nonlinear or requires multiple scenario evaluations. Greedy optimization over an approximate value function may also amplify errors, leading to unstable or suboptimal actions. 

Our work is also related to differentiable optimization layers and decision-focused learning. This line of work embeds an optimization problem into a learning pipeline and trains upstream models by differentiating, exactly or approximately, through the optimizer. Examples include differentiable convex or quadratic optimization layers, predict-then-optimize losses, MIP layers, black-box solver differentiation, and perturbed combinatorial optimizers \citep{elmachtoub2022smart,mandi2024decision,amos2017optnet,agrawal2019differentiable,ferber2020mipaal,vlastelica2020differentiation,berthet2020learning}. These methods are powerful when the downstream optimization problem is continuous or admits a useful differentiable surrogate, because they enable end-to-end training with gradients that directly reflect downstream decision quality. 

Bellman-Taylor score decoding utilizes optimization in a different way. The decoder is not a trainable differentiable layer, but a fixed action-selection map inside MDP. Given a latent score, it selects a feasible natural action over the original state-dependent feasible set. The score policy is trained from MDP return using likelihood-ratio policy gradients, so the optimizer is used only in the forward pass and need not be differentiated through. This distinction is important for operational MDPs with integer or combinatorial actions, where the exact decoder may be discontinuous and difficult to differentiate. 

Remarkably, \citet{hoppe2025structured} is also close to our approach, which maps the state to objective coefficients and uses a linear optimization to select the feasible action per iteration. Although the idea of algorithm design is similar, it requires a more specialized training pipeline. The discrete optimization layer is generally non-differentiable, so it relies on perturbation-based or differentiable surrogate losses for learning. As a result, their approach lacks theoretical performance guarantees and cannot be naturally integrated with off-the-shelf DRL algorithms, making it less straightforward to apply in practical operational MDPs.


\section{Problem Background} \label{background}

\subsection{Markov Decision Process}

Markov decision process (MDP) is a preeminent modeling paradigm in sequential decision-making under uncertainty \citep{sutton1998reinforcement}. Many applications in operations research (OR), including inventory management, queue scheduling, and resource allocation, can be naturally formulated as MDPs to solve. Throughout this paper, we focus on solving infinite-horizon, discounted-reward MDPs, while alternative formulations have been extensively studied in the literature \citep{puterman2014markov}.  

In general, a discounted-reward MDP is described by tuple $\mathcal{M} = (\mathcal{S}, \{\mathcal{A}(s)\}_{s \in \mathcal{S}}, \mathbb{P}, r, \gamma,\nu_0),$ where $\mathcal{S}$ is the state space and $\mathcal{A}(s)$ denotes the feasible action set at state $s$. Note that $\mathcal{A}(s)$ may vary for different states, i.e., it is state-dependent. $\mathbb{P}(\cdot \mid s, a)$ is the transition kernel, which governs the state transition's rule. Lastly, $r(s, a)$ is the one-period immediate reward, $\gamma \in (0,1)$ is the discount rate, and $\nu_0$ is  the distribution of initial state. A stationary policy $\pi$ samples an action $a_t \in \mathcal{A}(s)$ for each state $s_t$ in period $t$, following a fixed probability distribution $\pi(\cdot|s_t)$ supported on $\mathcal{A}(s_t)$. Thus, under policy $\pi$, starting from an state $s$, the expected accumulative discounted reward is $ V^\pi(s) = \mathbb{E}^\pi [ \sum_{t=0}^\infty \gamma^t r(s_t, a_t)| s_0=s ].$ The decision-maker's objective is to seek the optimal stationary policy to maximize the objective $ J(\pi;\nu_0)=\mathbb{E}_{s\sim \nu_0}[V^\pi(s)]\big\}$ over $\Pi$, the set of all stationary policies. Let $V^*(s) = \sup_{\pi \in \Pi} V^\pi(s)$ be the optimal value function. The corresponding optimal action-value function (or $Q$-function) is define as $ Q^*(s, a) = r(s, a) + \gamma \cdot \mathbb{E}_{s'\sim \mathbb{P}(\cdot|s,a)} \big[V^*(s') \mid s, a\big].$ 

According to dynamic programming principle, it is well-known that $V^*(s) = \max_{a \in \mathcal{A}(s)} Q^*(s, a)$ and the optimal policy $\pi^*$ chooses action in $\argmax_{a\in \mathcal{A}(s)}\{ Q^*(s, a) \}$ to take for each $s$. In addition, there exists a deterministic optimal policy that maximizes $J(\pi;\nu_0)$ \citep{sutton1998reinforcement}. 

It is useful to distinguish between MDPs with known and unknown transition dynamics. In the former case, the transition kernel $\mathbb{P}(\cdot|s,a)$ is explicitly specified, or at least a simulator is available to generate state transitions under any admissible action. In the latter case, the decision-maker only observes transition samples through interaction or from logged trajectories without knowing $\mathbb{P}(\cdot|s,a)$. In this paper, we focus on the known-transition (or simulator-available) setting, which is common in many OR applications where the system dynamics and operational constraints are explicitly specified.

\subsection{Applying DRL to OR: Challenges from State-Dependent Feasible Action Sets}

In many OR models, feasible action sets have complex state-dependent constraint geometry. Feasible actions are not drawn from a fixed finite action catalog, but are instead defined implicitly through operational constraints, integrality restrictions, and combinatorial coupling. For example, the feasible action set at state $s$ can be represented abstractly as
$$
\mathcal{A}(s)=\left\{a\in \mathcal{\bar{A}}: g_i(s,a)\le 0,\ i=1,\ldots,m_1,\ \ h_j(s,a)=0,\ j=1,\ldots,m_2 \right\},
$$
where \(\mathcal{\bar{A}}\) is an ambient mixed-integer action domain. The constraint functions encode the system's capacity limits, balance relations, and compatibility conditions. Furthermore, the action variables and feasible sets are often high-dimensional, coupled, or combinatorial. Even when individual action components are simple, the joint feasible set can be highly coupled and difficult to enumerate. The number of feasible actions often grows rapidly with system size, rendering explicit enumeration prohibitive.

These features do not align with standard action interfaces used by standard DRL implementations. Value-based methods require evaluating or maximizing over candidate actions, which is nontrivial when the feasible set cannot be explicitly enumerated. Policy-based methods typically parameterize distributions over simple action domains, such as fixed finite catalogs or Euclidean boxes, which is equally problematic when feasible actions form an irregularly constrained set. Mechanisms such as action masking is not a general remedy because they still rely on the existence of an tractable explicit finite superset of candidate actions, which can be prohibitively large. This limits the direct applicability of off-the-shelf DRL methods in OR.  

\section{Bellman-Taylor Score Decoding} \label{algorithm}

Many MDPs in OR applications admit a structural representation of the state transition. Throughout this paper, we investigate a particular class of problems in which, for each state $s$ and natural action $a \in \mathcal{A}(s)$, the next state $s'$ can be represented as $s' = \Xi_s(\phi_s(a), \xi_s)$. Here $\phi_s : \mathcal{A}(s) \to \mathcal{X}_s \subseteq \mathbb{R}^{d_s}$ represents a post-action configuration of system status, which captures an intermediate state immediately after the action is applied but before the uncertainty is realized. $\xi_s$ is an exogenous disturbance whose distribution is known and may depend on $s$, and $\Xi_s(\cdot, \cdot)$ is a transition function with known analytical formula. For notational consistency, we denote the one-period reward as $r(s, a) = \psi_s(a)$, where $\psi_s : \mathcal{A}(s) \to \mathbb{R}$ is a known state-dependent map that quantifies the immediate contribution of action $a$.

For instance, in a queueing-control problem, the state $s$ may describe the current queue lengths and server availability, while the natural action $a\in \mathcal{A}(s)$ specifies a dispatching or service-allocation decision. The post-action state $\phi_s(a)$ then denotes the queue configuration immediately after the action is implemented but before any new arrivals or service completions occur. The disturbance $\xi_s$ represents the exogenous randomness during the period, such as random arrivals or service outcomes, and $\Xi_s\bigl(\phi_s(a),\xi_s\bigr)$ maps the post-decision state and the realized randomness to the next state. In such problems, the optimal action-value function has additional structure beyond the generic MDP formulation, allowing for approximation via a Taylor expansion. This approximation then induces a score-based view of action selection.

\subsection{Motivation and Decoder-Induced Score Policy} \label{sec: motivation}

Given above structural transition representation, we define the continuation-value function (or the value-to-go function) of post-action configuration as
\begin{equation} \label{continuation-value function}
    G^*_s(x)=\mathbb{E}\!\left[V^*\bigl(\Xi_s(x,\xi_s)\bigr)\right], \ x\in \mathcal{X}_s,
\end{equation}
which denotes the expected optimal downstream reward when the system is placed at post-decision configuration $x$ and the subsequent evolution is driven by the stochastic disturbance $\xi_s$. Then the optimal $Q$-function admits decomposition $Q^*(s,a)=\psi_s(a)+\gamma\,G^*_s\bigl(\phi_s(a)\bigr), \ a\in \mathcal{A}(s)$.

Since the optimal action at state $s$ is $\argmax_{a\in \mathcal{A}(s)} Q^*(s,a)$, the difficulty to solve the MDP essentially lies in learning the continuation-value function $G^*_s(x)$. To obtain a tractable surrogate for $G^*_s\bigl(\phi_s(a)\bigr)$, given a state $s$, we choose a reference action $a^{\text{ref}}(s) \in \mathcal{A}(s)$ and  assume the corresponding post-action configuration is
$x^{\text{ref}}(s) = \phi_s(a^{\text{ref}}(s)) \in \mathcal{X}_s$.
We then approximate $G^*_s(\phi_s(a))$ locally via a first-order Taylor expansion at $x^{\text{ref}}(s)$, 
\begin{align} \label{first taylor}
    G^*_s(\phi_s(a)) \approx G^*_s(x^{\text{ref}}(s)) +\big \langle \nabla G^*_s(x^{\text{ref}}(s)), \phi_s(a) - x^{\text{ref}}(s) \big \rangle.
\end{align}
Although the state, action, and post-action configuration may be discrete, for analytical convenience, we still assume that $G^*_s(x)$ has a smooth extension around $x^{\text{ref}}(s)$. Thus, the gradient and Taylor expansion are well-defined. Substituting this expansion into the optimal $Q$-function gives
\begin{align*}
        Q^*(s, a) &\approx   \gamma G^*_s(x^{\text{ref}}(s))- \gamma \big \langle \nabla G^*_s(x^{\text{ref}}(s)), x^{\text{ref}}(s) \big \rangle + \psi_s(a)  +  \big \langle \gamma  \nabla G^*_s(x^{\text{ref}}(s)), \phi_s(a)  \big \rangle  \\
        &:= \alpha_1(s) + \psi_s(a) + \langle z_1^*(s), \phi_s(a) \rangle.    
\end{align*}
Since the intercept $\alpha_1(s)$ does not rely on $a$, the optimal action of state $s$ can be approximated as
\begin{align} \label{action approximation}
     \argmax_{a \in \mathcal{A}(s)} Q^*(s, a) \approx \argmax_{a \in \mathcal{A}(s)} \big\{ \psi_s(a) + \langle z_1^*(s), \phi_s(a) \rangle \big\}.
\end{align}

Equation \eqref{action approximation} suggests that the Bellman-greedy action ranking can be approximated by a score-based linear surrogate over the post-action configuration. This motivates us to restrict policy optimization in a class of decoder-induced policies, where the policy outputs a latent marginal-value score and the decoder maps this score to a feasible action by solving the surrogate maximization problem. Specifically, given a latent score $z \in \mathbb{R}^{d_s}$, define the Bellman–Taylor action decoder
\begin{align} \label{decoder def}
    \Gamma(s, z) \in \argmax_{a \in \mathcal{A}(s)} \big \{ \psi_s(a) + \langle z, \phi_s(a) \rangle \big\} \in \mathcal{A}(s).
\end{align}
We aim to optimize a stationary policy $\tilde{\pi}$ that maps $s$ to $z$, and then recovers the original natural action $a$ through \eqref{decoder def}. In this sense, the original MDP's action is transformed: the decision variable is no longer the natural action $a$ itself, but a score vector $z$ living in an Euclidean space $\mathbb{R}^{d_s}$ without any restrictions. 

Through this construction, action feasibility is enforced exactly by the decoder, where the MDP policy does not need to handle the irregular geometry of the original action space directly. As a result, policy optimization is moved to a fixed Euclidean score space. A standard continuous-action policy can generate latent scores, while the decoder converts each score into an implementable natural action before interacting with the original MDP.

We emphasize that the reference post-action configuration $x^{\text{ref}}(s)$ is used to motivate and analyze the Taylor expansion only. It does not enter the construction of decoder $\Gamma(s,z)$. Thus, our method can be implemented without specifying a $x^{\text{ref}}(s)$. Also note that optimizing the latent score policy is generally not equivalent to optimizing over all stationary policies of the original MDP, because the decoder-induced score policy space is restricted. Nevertheless, as long as the remainder terms of Taylor expansion \eqref{first taylor} can be controlled, the induced action ranking error can be controlled, leading to a bounded optimality gap. In Section \ref{sec:performance-analysis}, we make this statement precise. 

Lastly, we discuss some costs of this method. First, each action selection now requires solving the decoder optimization problem, which introduces additional computational overhead relative to directly parameterized policies. Second, the latent score MDP is only approximately aligned with the original optimization, since it is induced by an first-order Taylor approximation of the continuation-value function. The optimality gap depends on how accurately the induced surrogate preserves the Bellman action ranking. To mitigate these difficulties, note that many OR problems admit well-performed benchmark heuristics, which can be used to warm-start the DRL training via various techniques like behavior cloning. This may reduce the iteration numbers needed for training. In addition, to handle continuation-value functions with large curvature, we generalize above method via a higher-order Taylor expansion in Section \ref{sec: generalization}, yielding a richer decoder-induced score policy family that better aligns with the original problem.

\subsection{Solving Latent Score MDP via PPO} 

Let $\mathcal Z\subseteq \mathbb R^{d_s}$ be the latent score space. Given the
action decoder $\Gamma(s,z)$ \eqref{decoder def}, we define the decoder-induced MDP $\widetilde{\mathcal M} = (\mathcal S,\mathcal Z,\widetilde P,\widetilde r,\gamma,\nu_0)$, where the state space remains $\mathcal S$ and the action space is now $\mathcal Z$. For each state-score pair $(s,z)$, the decoder first
selects a feasible natural action $a=\Gamma(s,z)\in \mathcal{A}(s)$, and the induced reward and transition kernel are $\widetilde r(s,z)=r(s,\Gamma(s,z))$ and $\widetilde{\mathbb{P}}(\cdot\mid s,z)=\mathbb P(\cdot\mid s,\Gamma(s,z))$ respectively. If the decoder has multiple maximizers, we choose an arbitrary tie-breaking rule. A stationary score policy $\widetilde\pi$ samples
$z_t\sim \widetilde\pi(\cdot\mid s_t)$ first and executes the decoded natural action
$a_t=\Gamma(s_t,z_t)$. Its expected accumulative discounted value is $ \widetilde V^{\widetilde\pi}(s)
= \mathbb E^{\widetilde\pi}
[ \sum_{t=0}^{\infty}\gamma^t \widetilde r(s_t,z_t) | s_0=s].$

To optimize MDP $\widetilde{\mathcal{M}}$, we essentially need to solve a continuous-action control problem, where PPO is a standard approach. In PPO, the policy is parameterized as $\pi_\theta(z|s)$, where $\theta$ is optimized via gradient-based methods. Although the decoder $\Gamma(s,z)$ is an optimization program which can be nonsmooth in $z$, applying PPO to solve $\widetilde{\mathcal{M}}$ does not require differentiating through
$\Gamma(s,z)$. The policy gradient only involves $\nabla_\theta\log\pi_\theta(z\mid s)$
rather than derivative of action decoder. This is an important distinction from differentiable optimization-layer approaches. We call this procedure to solve the original MDP Bellman-Taylor Score Decoding PPO (BTSD-PPO) algorithm. 


Beyond PPO, the decoder-induced MDP $\widetilde{\mathcal M}$ can be solved by other DRL algorithms as well. Any method applicable for continuous-action control can be applied directly. Value-based methods such as DQN may also apply with an appropriate discretization of the latent score space to generate a finite action set. To be focused, the subsequent discussions are restricted to PPO. In the numerical study, we also test other DRL algorithms to demonstrate that the empirical benefit is primarily due to the Bellman-Taylor score decoding, rather than a particular optimizer.

\subsection{Generalization: Higher-Order Action Decoder} \label{sec: generalization}

The first-order Bellman-Taylor action decoder constructed above relies on a linear approximation of the continuation-value function, which is effective when the continuation effect is locally close to linear around the reference post-decision point. However, if the continuation-value function exhibits non-negligible curvature, then a first-order Taylor expansion may no longer capture the Bellman action ranking with sufficient accuracy. In such cases, it is natural to retain higher-order terms in the Taylor expansion, thereby enriching the features used by the action decoder.

We next introduce some notations to ease presentation. Fix an integer \(K\ge 1\), for a multi-index \(\alpha=(\alpha_1,\ldots,\alpha_{d_s})\in\mathbb{N}^{d_s}\), let $|\alpha|=\sum_{i=1}^{d_s}\alpha_i $ and $\alpha!:=\prod_{i=1}^{d_s}\alpha_i! $. For any \(x\in\mathcal X_s\), define
$$ \bigl(x-x^{\mathrm{ref}}(s)\bigr)^\alpha =
\prod_{i=1}^{d_s}\bigl([x]_i-[x^{\mathrm{ref}}(s)]_i\bigr)^{\alpha_i}=
\prod_{i=1}^{d_s}\bigl([x]_i-[\phi_{s}{(a^\mathrm{ref}(s))}]_i\bigr)^{\alpha_i},
$$
where $[a]_i$ denotes the $i$-th coordinate of a vector $a$. Under regularity conditions, the continuation-value function admits the $K$-th order Taylor expansion
\begin{equation} \label{high order expansion}
G^*_s\bigl(\phi_s(a)\bigr)
\approx
G^*_s\bigl(x^{\mathrm{ref}}(s)\bigr)
+
\sum_{1\le |\alpha|\le K}
\frac{D^\alpha G^*_s\bigl(x^{\mathrm{ref}}(s)\bigr)}{\alpha!}
\bigl(\phi_s(a)-x^{\mathrm{ref}}(s)\bigr)^\alpha,
\end{equation}
where $D^\alpha G^*_s(\cdot)$ denotes the $\alpha$-th order partial derivative of function $G^*_s(\cdot)$. If we define the lifted feature map
\begin{equation*}
F_{s,K}(a)
:=
\left(
\frac{\bigl(\phi_s(a)-x^{\mathrm{ref}}(s)\bigr)^\alpha}{\alpha!}
\right)_{1\le |\alpha|\le K},
\end{equation*}
whose dimension is $ m_{s,K}=\sum_{k=1}^{K}\binom{d_s+k-1}{k}$, 
and substitute the expansion \eqref{high order expansion} into the optimal \(Q\)-function, we obtain
\begin{equation*}
Q^*(s,a)
\approx
\alpha_K(s)+\psi_s(a)+\langle z_K^*(s),F_{s,K}(a)\rangle,
\end{equation*}
where \(\alpha_K(s)\) is an intercept term independent of \(a\), and
$ z_K^*(s) = \gamma (D^\alpha G^*_s(x^{\mathrm{ref}}(s)) )_{1\le |\alpha|\le K} \in \mathbb R^{m_{s,K}}.$
Hence, the optimal action corresponding to state \(s\) can be locally approximated by
\begin{equation*}
\argmax_{a\in \mathcal{A}(s)}Q^*(s,a)
\approx
\argmax_{a\in \mathcal{A}(s)}
\left\{
\psi_s(a)+\langle z_K^*(s),F_{s,K}(a)\rangle
\right\}.
\end{equation*}

As a result, under the \(K\)-th order Taylor expansion, the Bellman-greedy action ranking is determined by a higher-order score vector \(z_K^*(s)\). Instead of optimizing over the original natural action space, it suffices to learn the map from $s$ to \(z_K^*(s)\). For a latent score vector \(z\in\mathbb R^{m_{s,K}}\), define the higher-order Bellman-Taylor decoder
\begin{equation}
\Gamma_K(s,z)
=
\arg\max_{a\in \mathcal{A}}
\left\{
\psi_s(a)+\langle z,F_{s,K}(a)\rangle
\right\}\in \mathcal{A}(s).
\end{equation}
Then we obtain an optimization on the Euclidean space \(\mathbb R^{m_{s,K}}\) where $\Gamma_K(s,z)$ maps latent score to the natural action.  

The induced reparameterized MDP is defined analogously by replacing the first-order decoder \(\Gamma\) with \(\Gamma_K\). Specifically, for each state \(s\in\mathcal S\) and latent score \(z\in\mathbb R^{m_{s,K}}\), define
$ \tilde r_K(s,z):=r(s,\Gamma_K(s,z))$ and $
\tilde{\mathbb P}_K(\cdot\mid s,z):=\mathbb P(\cdot\mid s,\Gamma_K(s,z))$.
A latent-score policy on the corresponding Euclidean score space then induces a policy on the original natural-action space exactly as in the first-order case. All subsequent constructions, including the value function, the policy-gradient update, and the PPO implementation, extend verbatim after replacing \(\phi_s(a)\) with \(F_{s,K}(a)\) and \(\Gamma\) with \(\Gamma_K\).

In practice, we can reorganize the Taylor expansion \eqref{high order expansion} into an uncentered polynomial basis such that the constructed action decoder does not rely on reference action $x^{\mathrm{ref}}(s)$ explicitly. By the multi-index binomial identity,
\[
\bigl(\phi_s(a)-x^{\mathrm{ref}}(s)\bigr)^\alpha
=
\sum_{0\le \beta\le \alpha}
\binom{\alpha}{\beta}
\phi_s(a)^\beta
\bigl(-x^{\mathrm{ref}}(s)\bigr)^{\alpha-\beta},
\]
there exist a state-dependent intercept \(\widetilde\alpha_K(s)\) and a coefficient vector \(\widetilde z_K^*(s)\in\mathbb R^{m_{s,K}}\) such that
\[
Q^*(s,a)
\approx
\widetilde\alpha_K(s)+\psi_s(a)
+\big\langle \widetilde z_K^*(s),\widetilde F_{s,K}(a)\big\rangle,
\]
where the uncentered lifted feature map is defined by
$ \widetilde F_{s,K}(a)=({\phi_s(a)^\beta})_{1\le |\beta|\le K}.$
Accordingly, one may equivalently define the higher-order Bellman-Taylor decoder in the uncentered form
\[
\widetilde \Gamma_K(s,z)
=
\argmax_{a\in \mathcal{A}(s)}
\big\{
\psi_s(a)+\langle z,\widetilde F_{s,K}(a)\rangle
\big\}.
\]
This representation absorbs the reference point \(x^{\mathrm{ref}}(s)\) into the state-dependent coefficients and removes its explicit appearance from the decoder objective. Thus, same as the first-order method, we do not need to specify the choice of reference post-action configuration $x^{\text{ref}}(s)$ in algorithm. 

The higher-order action decoder therefore involves a clear tradeoff. Retaining more terms in the local Taylor expansion yields a richer surrogate for the continuation-value function, and can better preserve the Bellman action ranking when local curvature is large. However, this gain comes at a computational price. As the expansion order \(K\) increases, the lifted feature dimension $m_{s,K}$
grows rapidly, so the action space of the score-induced MDP is high-dimensional, the policy optimization becomes more expensive, and the action decoder must be evaluated on a larger feature set. In many OR problems of interest, however, the dominant structural effect is already captured by the first-order marginal continuation value, especially when the state transition exhibits certain affine structure. For this reason, the first-order decoder often provides a favorable balance between approximation quality, computational tractability, and ease of implementation, and will be our main focus in the numerical study.

\subsection{Performance Analysis}
\label{sec:performance-analysis}

In this section, we analyze the performance of Bellman-Taylor score decoding. To ease presentation, we focus on the first-order decoder and the analysis for higher order case is similar. Let $\hat{z}_1(s)$ be the policy trained by PPO to solve the decoder-induced MDP $\widetilde{\mathcal M}$. When a stochastic policy is adopted, $\hat{z}_1(s)$ becomes a random variable. $\hat{z}_1(s)$ induces a policy $ \hat{\pi}_1$ to the original MDP $\mathcal{M}$ that maps state $s$ to a natural action $a$ through the decoder $\hat{\pi}_1(s) = \Gamma(s, \hat{z}_1(s)).$ Let $\Pi_\Gamma$ be the decoder-induced policy class to MDP $\mathcal{M}$. Also define the best achievable value within the decoder-induced class by $J_\Gamma^*(\nu_0)
=\sup_{\pi\in\Pi_\Gamma} J(\pi;\nu_0).$ If the supremum is attained, we denote the maximizer by $\pi_\Gamma^*$, so that
$J_\Gamma^*(\nu_0)=J(\pi_\Gamma^*;\nu_0)$. Our objective is to bound  $J(\pi^*;\nu_0)-J_{\nu_0}(\hat{\pi}_1;\nu_0)$, where $\pi^*$ is the optimal policy to the original MDP $\mathcal{M}$. Throughout this section, we assume that all value functions are finite and that the Bellman maximizer and decoder maximizer admit measurable selections under the fixed tie-breaking rule used by the decoder. We fist impose an assumption to quantify the performance of PPO algorithm.

\begin{assumption}
\label{ass:first_order_learning_error}
There exists some constant $\mathcal E_{\rm learn}\ge 0$ such that $
J_\Gamma^*(\nu_0)-J(\hat\pi_1;\nu_0) \le \mathcal E_{\rm learn}.$
\end{assumption}

{Assumption~\ref{ass:first_order_learning_error} measures how well PPO solves the latent-score MDP relative to the best policy representable by the decoder-induced class. In practice, $\mathcal E_{\rm learn}$ includes finite-sample error, function-approximation error, and optimization error, etc. The convergence guarantee of PPO algorithm has been established in \cite{liu2019neural} and the references therein, which supports the rationale of this assumption.}

For any measurable deterministic score map \(z(\cdot):\mathcal{S}\to Z\), let \(\pi^z\) denote the deterministic decoded policy $\pi^z(s)=\Gamma(s,z(s)).$
For each state \(s\) and action $a$, define the Bellman residual after subtracting the linear score \(z\) by $R_s^z(a)=
\gamma G_s^*(\phi_s(a))-\langle z,\phi_s(a)\rangle$
and define its oscillation over the feasible action set $\mathcal{A}(s)$ by
\[
\Omega_s(z)=\operatorname{osc}_{a\in \mathcal{A}(s)}\big[R_s^z(a)\big]
=
\sup_{a\in \mathcal{A}(s)}R_s^z(a)-\inf_{a\in \mathcal{A}(s)}R_s^z(a).
\]
It easy to check that 
$ \Omega_s(z)= 2\inf_{c\in\mathbb R}
\sup_{a\in \mathcal{A}(s)}
\left|
\gamma G_s^*(\phi_s(a))-c-\langle z,\phi_s(a)\rangle
\right|$. Thus, \(\Omega_s(z)\) measures the maximal remaining variation in the Bellman continuation value at $s$ that cannot be explained by the linear score \(z\). We have the following theorem. 

\begin{theorem}
\label{thm:oscillation-bound}
Suppose Assumption~\ref{ass:first_order_learning_error} holds. Then
\[
J(\pi^*;\nu_0)-J(\widehat \pi_1;\nu_0)
\le
\inf_{z(\cdot):\mathcal{S}\to Z}
\left\{
\frac{1}{1-\gamma}
\mathbb E_{s\sim d_{\nu_0,\gamma}^{\pi^z}}
\left[
\Omega_s(z(s))
\right]
\right\}
+\mathcal{E}_{\mathrm{learn}}.
\]
\end{theorem}

Theorem~\ref{thm:oscillation-bound} decomposes the optimality gap of policy $\widehat \pi_1$ into two terms. The first term is a structural approximation error: it measures how much Bellman continuation-value variation remains after the best state-dependent linear score is removed. The second term is the algorithmic learning error incurred when PPO approximately solves the decoder-induced MDP.

The oscillation bound also gives a direct condition for exact Bellman action agreement. For each
state \(s\), define the Bellman action gap
\[
\Delta_s
=
V^*(s)
-
\sup_{a\in \mathcal{A}(s)}\big\{Q^*(s,a):\ Q^*(s,a)<V^*(s)\big\}.
\]
Fix a state \(s\) and a score \(z\in Z\). If \(\Omega_s(z)<\Delta_s\), then every action selected
by the decoder is Bellman optimal, i.e.,
$\Gamma(s,z)\in \arg\max_{a\in \mathcal{A}(s)}Q^*(s,a).$ This shows that an exact affine representation of \(G_s^*(\cdot)\) is not necessary for correct
action selection; it is enough for the residual oscillation to be smaller than the Bellman action gap.
Thus, the residual oscillation term in Theorem~\ref{thm:oscillation-bound} is a worst-case
certificate of statewise Bellman-ranking error; it may be nonzero without affecting action selection
whenever it remains below the local Bellman action gap.

Theorem \ref{thm:oscillation-bound} does not require a smooth and continuous extension of $G^*_s(\cdot)$ and therefore applies to finite, discrete, or continuous feasible action sets. However, when additional structure is available, the residual oscillation term can be characterized more explicitly as two examples below.

\textbf{Smooth Interpolant and Curvature-Radius Characterization.} For each state \(s\), let $\Phi_s=\{\phi_s(a):a\in \mathcal{A}(s)\}$ be the attainable post-action set. In a discrete MDP, the original model only defines the continuation value on \(\Phi_s\). Assume the continuation value function admits a low-curvature smooth interpolant.

\begin{assumption}
\label{ass:smooth-interpolant}
For each state \(s\), there exist an open convex set \(U_s\subseteq \mathbb R^{d_s}\) with $\operatorname{conv}(\Phi_s)\subseteq U_s,$ and a twice continuously differentiable function \(\bar G_s^* :U_s\to\mathbb R\) such that $\bar G_s^*(x)=G_s^*(x),\ \forall x\in \Phi_s$ 
and $\sup_{x\in \operatorname{conv}(\Phi_s)} \left\|
\nabla^2 \bar G_s^*(x)
\right\|_{\mathrm{op}}
\le L_2(s)..$
\end{assumption}

The smooth interpolant is an analytical device used to certify that Bellman continuation value function has bounded curvature on the relevant post-action geometry.  For any reference map \(x(\cdot)\) satisfying \(x(s)\in\operatorname{conv}(\Phi_s)\), define $z_x(s)=\gamma \nabla \bar G_s^*(x(s)).$ We have the following proposition. 

\begin{proposition}
\label{prop:smooth-curvature-bound}
Suppose Assumption~\ref{ass:smooth-interpolant} holds. Then $$\Omega_s(z_x(s)) \le \gamma L_2(s)
\sup_{a\in \mathcal{A}(s)} \|\phi_s(a)-x(s)\|_2^2. $$
\end{proposition}

Proposition~\ref{prop:smooth-curvature-bound} shows that the residual oscillation is a second-order local quantity. It is bounded by the local curvature of the
post-action continuation interpolant multiplied by the squared worst-case deviation of feasible post-action configurations from the reference point.

\textbf{Exact Finite-Action Characterization.} Theorem~\ref{thm:oscillation-bound} admits further simplification when feasible actions are finite for each state. Specifically, fix a state \(s\) and suppose $\mathcal A(s)=\{a_1,\ldots,a_n\}.$ We denote by $x_i=\phi_s(a_i)\in\mathbb R^{d_s}$ and $y_i=\gamma G_s^*(x_i)$. Let \(y=(y_1,\ldots,y_n)^\top\) and the augmented feature matrix
$B_s=((1,x^\top_1),\cdots,(1,x^\top_n))^\top\in \mathbb R^{n\times(d_s+1)}.$ We also define the optimal statewise
oscillation
$
\Omega_1^*(s)=
\inf_{z\in\mathbb R^{d_s}}
\operatorname{osc}_{i=1,\ldots,n}
[
y_i-z^\top x_i
].
$

\begin{proposition} Let \(\Delta_n\) be the probability simplex in \(\mathbb R^n\). Then 
\label{prop:chebyshev-characterization}
\begin{align*}
    \Omega_1^*(s)
&=
2\min_{\theta\in\mathbb R^{d_s+1}}
\|y-B_s\theta\|_\infty
=\sup
\Big\{
|p^\top y-q^\top y|:
p,q\in\Delta_n,\
\sum_{i=1}^n p_i x_i=\sum_{i=1}^n q_i x_i
\Big\}.
\end{align*}
\end{proposition}

Proposition~\ref{prop:chebyshev-characterization} shows that, in the finite-action case,
the statewise oscillation is the Chebyshev affine approximation error of the Bellman continuation trace on the attainable post-action points. Equivalently, if two randomized mixtures of feasible actions have the same average post-action configuration, then any first-order score assigns them the same average score;
any difference in their average Bellman continuation values is invisible to the first-order decoder. To connect it back to the performance bound, suppose there exists a measurable selector $z^{\mathrm{loc}}(s)\in
\argmin_{z\in\mathbb R^{d_s}}\Omega_s(z)$, and let \(\pi^{*}\) be the corresponding decoded policy
$ \pi^{*}(s)=\Gamma(s,z^{\mathrm{loc}}(s)).$
Then Theorem~\ref{thm:oscillation-bound} implies
\[
J(\pi^*;\nu_0)-J(\widehat\pi_1;\nu_0)
\le
\frac{1}{1-\gamma}
\mathbb E_{s\sim d_{\nu_0,\gamma}^{\pi^{\mathrm{loc}}}}
\big[
\Omega_1^*(s)
\big]
+
\mathcal E_{\mathrm{learn}}.
\]

Overall, the preceding results separate the optimality gap into a structural approximation term and an
algorithmic learning term, and provide analytical characterizations of the structural term through
residual oscillation. However, these
bounds involves quantities that are defined through the optimal continuation value, which is itself the
solution of the Bellman fixed-point equation. For a general MDP, converting these quantities into explicit bounds involving only primitive model parameters is typically difficult: it requires controlling the geometry, curvature, or finite-difference structure of the optimal value
function over the relevant post-action configurations.

\section{Sanity Check: Application to Inventory Control} \label{inventory}
In this section, we consider a multi-location inventory control problem with cross-transshipment as a controlled diagnostic application of our method. The model is intentionally small and simplified, rather than a practical inventory operations. Its purpose is to demonstrate that the proposed action reparameterization framework can handle OR problems with state-dependent feasible action sets. 

Specifically, let $\mathcal I = \{1,\dots,n\}$ index inventory locations. At the beginning of each period, the on-hand inventory level of location $i$ is $s_i\in \{1,2,\cdots, L_i\}$ where $L_i$ denotes an inventory upper bound imposed for computational purposes so that the resulting MDP has a finite state space. Thus, $s = (s_1,\dots,s_n)$ denotes the system's state. Each location independently faces stochastic demand $d_i$ whose distribution is $\mathcal{D}_i$. To fill the demand, location $i$ orders a quantity of $u_i\in \{0,1,\cdots,U_i\}$ items to replenish inventory at the end of period. In addition to self-replenishment, transshipment across different locations is allowed. Let $y_{ij} \in \mathbb Z_+$ denote the quantity transshipped from location $i$ to $j$. Thus, the action of this problem is $a=(y,u)$. Since the total transshipment from location $i$ cannot exceed its current inventory capacity $s_i$, the feasible action space of state $s$ is 
$$ \mathcal A(s) = \Big\{(y,u): y_{ij} \in \mathbb Z_+,\ u_i \in \{0,\dots,U_i\},\ \sum_{j\neq i} y_{ij} \le s_i,\ \forall i \in \mathcal I \Big\}.$$ Accordingly, we define outbound and inbound transshipment of location $i$ as $o_i(y) = \sum_{j\neq i} y_{ij}$ and $m_i(y) = \sum_{j\neq i} y_{ji}$. 

{To increase the nonlinearity of system dynamics, we assume that cross-transshipment may incur losses, so not all inbound shipments are successfully processed. This captures the effect of receiving congestion and diminishing marginal contribution of large inbound volumes. Specifically, given an inbound volume $m_i(y)$, the successfully received inbound inventory of location $i$ is a random variable $e_i \sim \mathrm{Binomial}(m_i(y), p_{i,\rho}(m_i(y))$, where the success probability satisfies $p_{i,\rho}(m) = 1 - \rho {(m-1)^+}/{\kappa_i}$ where $\kappa_i = \max\{1,  \sum_{j\neq i} L_j - 1\}$. Note that $0\le \rho <1$ is a parameter that controls the strength of receiving cost in cross-transshipment, with larger $\rho$ reducing the probability that inbound inventory is successfully processed and incorporated into the available stock. When $\rho=0$, all inbound transshipment is received. After receiving self-replenishment and cross-transshipment, demand $d=(d_1,\dots,d_n)$ is realized and the next-period inventory level of location $i$ is updated as $s_i' = \min\{L_i, (s_i-o_i(y)+e_i+u_i-d_i)^+\}.$
Let $r_i=s_i-o_i(y)+e_i+u_i$ be the total inventory level of location $i$ immediately before $d_i$ is realized. Then the one-period cost is
\[
c(s,a) = \sum_{i\neq j} \tau_{ij} y_{ij} + \sum_i \chi_i u_i + \sum_i h_i (r_i - d_i)^+ + \sum_i \ell_i (d_i - r_i)^+,
\]
where $\tau_{ij},\chi_i,h_i,\ell_i$ are the unit transshipment, replenishment, holding, and lost-sale costs, respectively. The objective is to minimize  $J(\pi;\nu_0) = \mathbb E_{s_0\sim\nu_0}^\pi [ \sum_{t=0}^\infty \gamma^t c(s_t,a_t)]$ where $\nu_0$ is the initial distribution. 

In this formulation, the post-action configuration is $\phi_s(a) = (b,m) \in \mathbb R^{2n}$, where $b_i = s_i - o_i(y) + u_i$ is the local inventory base after outbound transshipment and replenishment, and $m_i = m_i(y)$ is the nominal inbound transshipment volume. The randomness of system transition are incorporated by the received inbound transshipment and demand, which can be represented as $\Xi_s(\phi_s(a), \xi_s)$. Thus, this problem can be solved by our proposed method. We test the first-order and second-order action decoder, i.e., 
$ \Gamma_K(s,z) \in \argmin_{a \in \mathcal A(s)} \{ c(s,a) +  \langle z, F_K(\phi_s(a)) \rangle \}$, where $F_1(w)=w$ is an identity map and $F_2(w)=(w,w\otimes w)$ incorporates all second-order terms. 

Let $\widehat \pi_{K}$ be the optimal policy trained by PPO with $K$-th order action decoder. For each problem instance, we also exactly calculate the optimal policy $\pi^*$ value as benchmark, which is achievable for small-scale problems. The corresponding value function and $Q$-function are $V^*(s)$ and $Q^*(s,a)$. We compare $\widehat \pi_{K}$ with the $\pi^*$ via three metrics: (1) optimality gap $J(\widehat \pi_{K};\nu_0)/J(\pi^*;\nu_0)-1$; (2) optimal action agreement rate; and (3) Bellman regret $\mathbb E_{x\sim\mu_\rho^*}
[Q_\rho^*(x,a_{\rho,K}^{\rm br}(x))-V_\rho^*(x)],$ where $\mu_\rho^*$ is the occupancy measure induced by $\pi^*$. We test 4 instances where parameter $\rho$ varies in $\{0,0.25,0.5,0.75\}$. {As $\rho$ increases, the post-action transition becomes more nonlinear, in the sense that the successfully received inbound inventory exhibits stronger diminishing marginal returns with respect to the nominal inbound transshipment volume.} Other model parameters, as well as PPO training details, are provided in Appendix. The experiment results are summarized in Table \ref{tab:inventory_policy}.}

\begin{table}[t]
\centering
\caption{\textbf{Performance for first-order and second-order decoded policies in the inventory model.}}
\label{tab:inventory_policy}
\small
\setlength{\tabcolsep}{5pt}
\renewcommand{\arraystretch}{1.05}
\begin{tabular*}{0.96\textwidth}{@{\extracolsep{\fill}}c ccc ccc@{}}
\toprule
& \multicolumn{3}{c}{First-order decoded policy}
& \multicolumn{3}{c}{Second-order decoded policy} \\
\cmidrule(lr){2-4}\cmidrule(lr){5-7}
$\rho$
& optimality gap & agreement & regret
& optimality gap & agreement & regret \\
\midrule
0
& 1.0\% & 95.7\% & 0.38
& 0.2\% & 98.8\% & 0.03 \\
0.25
& 1.5\% & 93.5\% & 0.60
& 0.3\% & 97.6\% & 0.08 \\
0.50
& 10.5\% & 51.1\% & 4.23
& 0.4\% & 96.1\% & 0.12 \\
0.75
& 11.4\% & 49.6\% & 5.08
& 0.6\% & 91.7\% & 0.27 \\
\bottomrule
\end{tabular*}
\normalsize
\end{table}
The results in Table~\ref{tab:inventory_policy} demonstrates the role of action decoder. The first-order decoder already yields effective decoded policies under the state-dependent transshipment constraints, particularly when the receiving process is close to linear. As the congestion parameter $\rho$ increases, the expected received quantity becomes a more curved function of the nominal inbound volume, and the performance of the first-order decoder gradually deteriorates. This is reflected in the increase of its optimality gap and Bellman regret, as well as the decline in the Bellman action agreement.

The second-order decoder mitigates this deterioration by allowing the score function to depend on quadratic features of the post-action configuration. Its optimality gap remains below $0.59\%$ and its Bellman regret remains small across all values of $\rho$, suggesting that the additional second-order terms help preserve the Bellman action ranking when nonlinear receiving effects are present. These findings highlight the intended tradeoff of the proposed framework: a first-order decoder offers a simple and computationally attractive reparameterization when the dynamics are nearly linear, whereas higher-order decoders provide additional flexibility for systems with complex transition mechanisms such as congestion, loss, saturation, or diminishing marginal processing efficiency.

\section{Case Study: Application to Queueing Network Control} \label{queue}

{Dynamic routing in multi-class and multi-pool service systems is a canonical problem of queueing network control, where heterogeneous customer classes need to be dynamically matched to server pools with different skills or service efficiencies  \citep{chen2020survey}. The routing decisions must balance several competing effects: assigning a customer to a flexible or non-primary pool may reduce current congestion, but it can also consume scarce capacity, lower service efficiency, or incur overflow costs. This model finds many applications in service systems such as inpatient overflow management \citep{dai2019inpatient,sun2024inpatient}. This problem can be naturally formulated as an MDP and the action space is state dependent: each action is an integer dispatching matrix whose feasibility is determined by the current queues and server capacities. As a result, this problems becomes a suitable testbed for our algorithm.}

\subsection{Model Description and MDP Formulation}

We consider a discrete-time, non-preemptive, parallel-service system with multiple customer classes and multiple server pools following the setup in \cite{chen2020survey}. Let
$\mathcal I=\{1,\ldots,I\}$ and $\mathcal J=\{1,\ldots,J\}$
denote the sets of customer classes and server pools, respectively. Customers of class \(i\in\mathcal I\) arrive according to a homogeneous Poisson process with rate \(\lambda_i\). Server pool \(j\in\mathcal J\) has capacity \(U_j\), meaning that at most \(U_j\) customers can be simultaneously served in pool \(j\). The service time of a class-$i$ customer served by a type-$j$ server is an exponential $\mu_{ij}$ distribution. Equivalently, in the discrete-time system, the probability of service completion in one time period is \(1-e^{-\mu_{ij}}\in (0,1)\). Let
$ \mathcal E=\{(i,j)\in\mathcal I\times \mathcal J:\mu_{ij}>0\}$ be the set of all compatible class-pool pairs. This setup generalizes the pool-dependent service rate model studied in \citep{dai2019inpatient,sun2024inpatient}, but also increases the state's dimension.

At the beginning of each period \(t\), the system state is $s_t=(q_t,h_t)$, where
$q_t=(q_{1,t},\ldots,q_{I,t})\in \mathbb Z_+^I$ records the queue lengths, and
$h_t=(h_{ij,t})_{(i,j)\in\mathcal E}\in \mathbb Z_+^{|\mathcal E|}$ records the in-service occupancy by origin, with \(h_{ij,t}\) denoting the number of class-\(i\) customers currently being served in pool \(j\). This finer state description is necessary because, once the service rate depends on \((i,j)\), the total occupancy of pool \(j\) alone is no longer sufficient to characterize the departure process. The action of period \(t\) is a matrix $a_t=(a_{ij,t})_{(i,j)\in\mathcal E}$, where \(a_{ij,t}\in\mathbb Z_+\) denotes the number of class \(i\) customers dispatched from the queue to pool \(j\) at time \(t\). A feasible action must satisfy constraints $\sum_{j:(i,j)\in\mathcal E} a_{ij,t}\le q_{i,t}, \forall i\in\mathcal I$, which means that one cannot dispatch more class \(i\) customers than are currently waiting in queue \(i\), and $\sum_{i:(i,j)\in\mathcal E}\bigl(h_{ij,t}+a_{ij,t}\bigr)\le U_j,\forall j\in\mathcal J$, which means that the total occupancy of pool \(j\) after dispatching cannot exceed its service capacity. As a result, the feasible action set at state \(s=(q,h)\) is characterized by 
\begin{align} \label{action space def}
    \mathcal{A}(s)=
\Big\{
a=(a_{ij})_{(i,j)\in\mathcal E}\in\mathbb Z_+^{|\mathcal E|}
:
\sum_{j:(i,j)\in\mathcal E} a_{ij}\le q_i,\ 
\sum_{i:(i,j)\in\mathcal E}(h_{ij}+a_{ij})\le U_j
\Big\}.
\end{align}
Given state \(s_t=(q_t,h_t)\) and action \(a_t\), the dispatching is implemented first. Then random service completions and exogenous arrivals occur. Specifically, for each class \(i\), the queue lengths becomes $q_{i,t+1}= q_{i,t} -
\sum_{j:(i,j)\in\mathcal E} a_{ij,t} +\delta_{i,t},$ where $\delta_{i,t}\sim \mathrm{Poisson}(\lambda_i)$. For each compatible pair \((i,j)\in\mathcal E\), the in-service occupancy satisfies $h_{ij,t+1} = h_{ij,t} +a_{ij,t}-d_{ij,t},$ where $d_{ij,t}\sim \mathrm{Binomial}(h_{ij,t}+a_{ij,t},1-e^{-\mu_{ij}}).$

We consider a one-period cost consisting of queue holding cost and dispatching cost. Let \(w_i>0\) be the waiting-cost coefficient of class \(i\), and let \(c_{ij}\ge 0\) denote the one-time cost of dispatching a class \(i\) customer to pool \(j\). This allows \(c_{ij}\) to encode, for example, overflow penalties for sending customers to non-primary pools. Thus, the one-period cost is $ c(s_t,a_t)
=
\sum_{i=1}^I
w_i q_{i,t}
+
\sum_{(i,j)\in\mathcal E} c_{ij}a_{ij,t}$. 

A stationary dispatching policy $\pi$ specifies, for each state $s$, a probability distribution supported on the feasible action set $\mathcal{A}(s)$. Starting from an initial state $s$ whose distribution is $\nu_0$, the decision-maker's objective is to minimize the expected cumulative discounted cost
$
J(\pi;\nu_0) = \mathbb{E}_{s_0\sim \nu_0}^\pi[ \sum_{t=0}^\infty \gamma^t c(s_t, a_t)] ,
$
where $\gamma \in (0, 1)$ is the discount factor (or equivalently, maximize the negative accumulative cost as reward.)

\subsection{First-Order Score Decoding: Learning State-Dependent Index Rule}

Above queueing model fits the structural MDP studied in this paper. Given a state \(s=(q,h)\) and a feasible action \(a\in \mathcal{A}(s)\), define the post-action configuration $\phi_s(a)=(q^+(a),h^+(a))$, where
$q_i^+(a)=q_i-\sum_{j:(i,j)\in\mathcal E} a_{ij}$ and $h_{ij}^+(a)=h_{ij}+a_{ij}$. It represents the system configuration immediately after dispatching and before random arrivals or service completions occur. Also define the exogenous disturbance as $\xi_s=(\delta,d),$
where \(\delta=(\delta_i)_{i\in\mathcal I}\) is the new arrivals with independent components $\delta_i\sim \mathrm{Poisson}(\lambda_i)$, and \(d=(d_{ij})_{(i,j)\in\mathcal E}\) is the service-completion vector satisfying $d_{ij}\sim \mathrm{Binomial}(h_{ij}^+(a),1-e^{-\mu_{ij}}).$
Then the state transition can be written as $ s'=\Xi_s(\phi_s(a),\xi_s),$ where
$\Xi_s((q^+,h^+),(\delta,d)) = (q^+ + \delta,\ h^+ - d).$ The one period reward is also a function $\psi_s(a)= -\sum_{i=1}^I
w_iq_i - \sum_{(i,j)\in\mathcal E} c_{ij}a_{ij}.$

For the first-order Bellman-Taylor score decoding strategy introduced before, let $z=(z^q,z^h)\in\mathbb R^{I+|\mathcal E|},$
where \(z^q=(z_i^q)_{i\in\mathcal I}\) and \(z^h=(z_{ij}^h)_{(i,j)\in\mathcal E}\) are the latent scores associated with the post-decision queue lengths $q^+(a)$ and in-service occupancies $h^+(a)$, respectively. The action decoder is  
\begin{align*}
\psi_s(a)+\langle z,\phi_s(a)\rangle =
\mathrm{const}(s,z)
+
\sum_{(i,j)\in\mathcal E}
\bigl(
-c_{ij}-z_i^q+z_{ij}^h
\bigr)\cdot a_{ij}
\end{align*}
where \(\mathrm{const}(s,z)\) denotes an intercept term not depending on \(a\). Hence, in this queueing application, the first-order action decoder reduces to an integer program over the feasible dispatching set of format 
$ \Gamma(s,z)=
\argmax_{a\in \mathcal{A}(s)}
\sum_{(i,j)\in\mathcal E}
\tilde{z}_{ij}(s)\cdot a_{ij}$
with $\tilde{z}_{ij}(s)=-c_{ij}-z_i^q+z_{ij}^h$. With such a reparameterization, the formal queueing problem is translated into a continuous-action MDP on the Euclidean score space $\mathbb{R}^{I+|\mathcal{E}|}$. At each state $s$, the policy no longer generates the dispatching matrix $a$ directly, but instead, produces a state-dependent score vector $z$. 

This separation is crucial in queueing network control. If we parameterize a policy directly on the dispatching matrix $a$, the neural network must operate over a state-dependent action set defined by integer, queue-availability, and pool-capacity constraints specified in \eqref{action space def}, which is highly nontrivial and typically requires either explicit projection or ad hoc feasibility corrections. By contrast, after reparameterization, the policy acts on the latent score space $\mathbb{R}^{I+|\mathcal{E}|}$, which is a fixed Euclidean space independent of system status. The difficult combinatorial part of the decision is then delegated to the decoder, which maps the score vector to a feasible dispatching action by solving the induced integer program. Hence feasibility is enforced exactly by the decoder, while policy optimization is carried out on a regular latent space rather than on the original irregular action set. As a result, the queueing-control problem is converted into a latent-score MDP that can be solved using the standard PPO procedure efficiently.

This approach also finds connections with several classical routing rules from the queueing literature. With policy optimization, the decoder actually assigns to each compatible class-pool pair $(i, j)$ an effective state-dependent index $\tilde{z}_{ij}(s)$ and then chooses the feasible dispatching matrix that maximizes the total weighted sum $\sum_{(i,j) \in \mathcal{E}} \tilde{z}_{ij}(s)\cdot a_{ij}$. In this sense, the resulting policy is a learned max-weight-type dispatching rule. Classical heuristics such as the generalized $c\mu$ rule, max-weight rule \citep{BuyukkocVaraiyaWalrand1985,tassiulas1990stability,stolyar2004maxweight}, and their overflow-cost-adjusted variants \citep{chen2020survey} can all be viewed as hand-crafted specifications of such pairwise indices, typically expressed analytically in terms of queue lengths, holding costs, service rates, and overflow costs. Under our notation, $\tilde{z}_{ij}(s)$ can be $w_i\mu_{ij}$, $w_i\mu_{ij}-c_{ij}$, $ w_iq_i\mu_{ij}$, and $ w_iq_i\mu_{ij}-c_{ij}$, corresponding to the $c\mu$ rule, modified $c\mu$ rule, max-weight rule, and modified max-weight rule in literature. By contrast, our formulation does not fix the index $\tilde{z}_{ij}(s)$ a priori but learns the optimal through DRL automatically.

\subsection{Numerical Experiments}

We test our algorithm empirically in this section. Due to space limitation, we only report the experiment results here. 

\subsubsection{Small-scale setting.}
We first conduct a small-scale experiment in which the optimal dynamic policy can be computed exactly. This serves as a sanity check to verify how our algorithm performs against the optimal in theory. Particularly, we consider a \(2\times 2\) queueing network with two customer classes and two server pools. Each server pool has capacity 5, and all compatible class-pool pairs have homogeneous service-completion probability
\(\mu_{ij}=1\). Customer class $i$'s primary server pool is $i$, $i=1,2$. The waiting-cost coefficients are normalized to one. Dispatching a customer to its primary pool is cost-free, whereas to a non-primary pool incurs a routing cost. We consider three
overflow-cost levels \(c^{\mathrm{ov}}\in\{0.1,0.3,0.6\}\), and set $c_{ij}=c^{\mathrm{ov}}$ for $i\ne j$.

To simplify model, we assume the service rates only depend on server pool type, i.e., $\mu_{ij}=\mu_j$. We study two traffic configurations. In the balanced-load instance, both classes have utilization levels \(\rho_1=\rho_2=0.9\), where $\rho_i=\lambda_i/(U_i \mu_i)$. In the imbalanced-load
instance, the two classes have utilization levels
\(\rho_1=0.95\) and \(\rho_2=0.85\), which creates asymmetric congestion
pressure and makes routing decisions more state dependent. The system starts from empty. For each instance, we compute the optimal value function exact dynamic-programming solution on the
finite-state system, with an appropriate state truncation mechanism, to obtain the optimal value function \(V^*\). We then train the BTSD-PPO algorithm and evaluate its performance via simulation. {The results are summarized in Table \ref{tab:small_scale_scratch_tuned}. The number in bracket denotes the estimation standard error of policy evaluation via simulation. We note that BTSD-PPO achieves performance close to the exact optimal, with relative gaps below $2.5\%$ across all tested instances. This suggests that the first-order decoder-induced policy family indeed preserves the essential structure of the optimal policy in small systems where exact comparison is possible.}
\begin{table}[t]
\centering
\small
\setlength{\tabcolsep}{4pt}
\caption{\textbf{Comparison of BTSD-PPO with the exact optimal solution in a $2\times2$ network.}}
\label{tab:small_scale_scratch_tuned}
\begin{tabular}{lcccccc}
\toprule
& \multicolumn{3}{c}{$\rho=(0.90,0.90)$}
& \multicolumn{3}{c}{$\rho=(0.95,0.85)$} \\
\cmidrule(lr){2-4}\cmidrule(lr){5-7}
Metric
& $c^{\mathrm{ov}}=0.1$
& $c^{\mathrm{ov}}=0.3$
& $c^{\mathrm{ov}}=0.6$
& $c^{\mathrm{ov}}=0.1$
& $c^{\mathrm{ov}}=0.3$
& $c^{\mathrm{ov}}=0.6$ \\
\midrule
Exact optimal
& 1210.4\,(1.8)
& 1241.1\,(2.0)
& 1285.3\,(2.1)
& 1185.4\,(2.0)
& 1218.1\,(2.0)
& 1264.5\,(2.0) \\

BTSD-PPO
& \textbf{1226.3\,(1.8)}
& \textbf{1261.0\,(2.2)}
& \textbf{1316.9\,(2.1)}
& \textbf{1198.1\,(1.6)}
& \textbf{1234.2\,(1.9)}
& \textbf{1278.2\,(1.9)} \\

Gap 
& 1.3\%
& 1.6\%
& 2.5\%
& 1.1\%
& 1.3\%
& 1.1\% \\
\bottomrule
\end{tabular}
\end{table}
\subsubsection{Moderate-scale setting.}
We next consider a moderate-scale queueing network with \(I=5\) customer classes and \(J=5\) server pools. Each pool has capacity \(U_j=30\). The
network is fully connected, i.e., $ \mathcal E=\mathcal I\times \mathcal J.$ Customer class
\(i\) has primary pool \(i\). Thus, the action at each state is a
\(5\times 5\) integer dispatching matrix subject to the queue-availability and pool-capacity constraints in \eqref{action space def}. Same as above, service rates are pool dependent but class
independent. Specifically, for each server pool \(j\), we randomly select $ \mu_j\sim \mathrm{Unif}(0.75,0.95),$ and set $\mu_{ij}=\mu_j$. This specification preserves service-rate heterogeneity across server pools while keeping the moderate-scale setting comparable to classical pool-dependent service models. Problem of similar structure and scale has been investigated in \cite{dai2019inpatient} in the context of inpatient ward management. 

We consider three traffic-intensity regimes. Let $m_i=U_{i}\mu_{i} $
be the nominal primary-pool service capacity for class \(i\). For each regime \(r\), we specify a primary-load vector
\(\pmb{\rho}^{\,r}=(\rho_1^r,\ldots,\rho_5^r)\) and determine the arrival of $\lambda_i$ as $\lambda_i^r=\rho_i^r m_i$, $1\le i\le 5$. The first regime is a symmetric heavy-loaded system, where all classes have same utilization rate \(0.95\), i.e., $\pmb{\rho}^{1}=
(0.95,0.95,0.95,0.95,0.95).$ In the second regime, two classes are close to heavy traffic, while the remaining three classes are moderately loaded: $ \pmb{\rho}^{2} = (0.98,0.96,0.88,0.84,0.84).$ In the last regime, one class is moderately overloaded,
two classes are near heavy traffic, and two classes are relatively lightly loaded: $\pmb{\rho}^{3} = (1.05,0.96,0.92,0.80,0.77).$
Note that both imbalanced regimes have average nominal primary utilization equal to \(0.90\), but they create different congestion patterns. For each regime, we create three instances, by choosing the waiting cost $w_i$ and overflow cost $c_{ij}$ from $[0,10]$ such that optimal OR heuristic introduced below varies. 


{For benchmarks, we first consider four classical index-based routing heuristics discussed in \cite{chen2020survey}, namely, $c\mu$ rule, modified $c\mu$ rule, max-weight rule, and modified max-weight rule. These benchmarks are state-of-the-art OR heuristics literature and allow us to assess whether learning
state-dependent routing scores improves upon standard hand-crafted
indices. We also test two RL-based benchmark. The first is the PPO algorithm proposed in \cite{sun2024inpatient}, which decomposes system-level combinatorial actions into sequential atomic actions within a PPO framework, enabling scalable policy learning for large inpatient overflow problem. Another one is the MIP-based value-lookahead algorithm proposed in \cite{harsha2025deep}, which combines value-function approximation with mathematical programming to optimize per-step combinatorial actions for long-term rewards in multi-node inventory systems. We emphasize that these benchmarks are adapted implementations rather than direct replications of the original algorithms. The original methods were developed for different operational models and include problem-specific tailoring to those settings. For example, \citet{sun2024inpatient} investigate a time-varying Poisson arrival process while in our model, the arrival rate is fixed. These model-specific components are not directly applicable to our setting. The comparison of different methods of each problem setting is reported in Table \ref{tab:fixed_5x5_metric1}.}
\begin{table}[t]
\centering
\caption{\textbf{Comparison of BTSD-PPO with various benchmarks in a $5\times5$ network.}}
\label{tab:fixed_5x5_metric1}
\resizebox{\textwidth}{!}{%
\begin{tabular}{lcccccccc}
\toprule
Case & $c\mu$ & mod.\ $c\mu$ & m.p. & mod.\ m.p. & Atom-PPO & MIP & BTSD-PPO & Improve \\
\midrule
B1 & 233.6\,(1.2) & 373.2\,(1.2) & 310.2\,(0.4) & 310.0\,(0.1) & 297.6\,(0.2) & 323.3\,(0.8) & \textbf{224.2\,(0.2)} & +4.0\% \\
B2 & 459.2\,(1.3) & 373.8\,(0.8) & 550.8\,(0.7) & 501.5\,(0.3) & 418.1\,(0.7) & 373.0\,(0.5) & \textbf{351.2\,(0.9)} & +6.0\% \\
B3 & 954.0\,(1.6) & 903.8\,(1.2) & 999.9\,(3.0) & 583.0\,(2.0) & 677.5\,(1.3) & 904.3\,(2.3) & \textbf{547.8\,(1.2)} & +6.0\% \\
T1 & 225.1\,(2.7) & 368.7\,(1.1) & 305.4\,(0.6) & 302.8\,(1.3) & 289.5\,(0.3) & 367.8\,(1.3) & \textbf{208.5\,(1.5)} & +7.4\% \\
T2 & 468.4\,(1.6) & 368.2\,(1.0) & 545.5\,(1.7) & 492.0\,(1.2) & 606.8\,(2.6) & 370.0\,(1.3) & \textbf{346.4\,(0.4)} & +5.9\% \\
T3 & 545.7\,(0.6) & 600.5\,(1.2) & 610.3\,(1.3) & 518.3\,(1.1) & 758.4\,(13.1) & 595.2\,(0.9) & \textbf{395.7\,(1.7)} & +23.7\% \\
O1 & 233.1\,(0.9) & 359.0\,(2.3) & 305.2\,(1.3) & 302.2\,(1.5) & 349.4\,(0.8) & 272.3\,(0.5) & \textbf{196.5\,(3.8)} & +15.7\% \\
O2 & 474.8\,(0.9) & 365.6\,(1.3) & 544.6\,(1.1) & 494.8\,(1.6) & 499.6\,(0.7) & 365.3\,(1.5) & \textbf{305.0\,(1.0)} & +16.6\% \\
O3 & 487.7\,(1.3) & 522.9\,(1.5) & 598.5\,(1.9) & 491.3\,(1.2) & 574.7\,(3.8) & 522.7\,(1.7) & \textbf{414.9\,(2.9)} & +14.9\% \\
\bottomrule
\end{tabular}%
}
\end{table}

We observe that across all the nine cases, BTSD-PPO achieves the lowest cost, with improvements over the best benchmark ranging from 4.0\% to 23.7\%. We also note that the best-performing OR heuristic varies across instances, highlighting that fixed heuristics are sensitive to traffic patterns and cost structures. For the Atom-PPO baseline, it generates feasible actions sequentially and preserves feasibility but fail to capture the global coupling of queue lengths and pool capacities, which are consistently outperformed by BTSD-PPO. Lastly, While the MIP-based baseline can be strong in some cases, it lacks robustness across all scenarios. 

We further conduct an ablation study to isolate the role of BTSD from the choice of the underlying DRL optimizer. Specifically, we combine the same BTSD decoder with three representative DRL backbones: PPO, SAC, and DQN. We compare them with their vanilla counterparts that operate on the original action representation. A naive action masking and projection mechanism is used to guarantee the feasibility 
The results, presented in Table~\ref{tab:moderate_ablation_current}, show that BTSD consistently improves all three backbones. Relative to the vanilla implementations, BTSD reduces the cost by 44.1\% for PPO, 44.8\% for SAC, and 41.8\% for DQN. Moreover, all BTSD-enhanced variants outperform the best benchmark policy. In contrast, the vanilla DRL algorithms perform substantially worse than the heuristic benchmark. These results suggest that the performance gain is not driven by a particular DRL optimizer, but by the BTSD action interface itself, which converts the original state-dependent and combinatorial action space into a tractable latent-score decision space while preserving exact feasibility.

\begin{table}[t]
\centering
\caption{\textbf{Ablation Study: BTSD consistently improves different DRL backbones.}}
\label{tab:moderate_ablation_current}
\begin{tabular}{lccc}
\toprule
DRL Backbone & Vanilla & BTSD & Improvement \\
\midrule
PPO & 397.2\,(0.00) & \textbf{222.2}\,(0.02) & +44.1\% \\
SAC & 417.0\,(0.01) & \textbf{230.2}\,(0.22) & +44.8\% \\
DQN & 393.1\,(0.02) & \textbf{228.9}\,(0.04) & +41.8\% \\
\midrule
Best benchmark ($c\mu$-rule) & \multicolumn{2}{c}{234.3\,(0.05)} & -- \\
\bottomrule
\end{tabular}
\end{table}

\section{Conclusions and Future Work}
\label{conclusion}

This paper proposes Bellman-Taylor score decoding as an action interface for MDPs with state-dependent feasible action sets. The main idea is to separate learning from feasibility. A standard continuous-action DRL algorithm learns state-dependent scores in a Euclidean latent space, while an optimization decoder maps these scores into feasible natural actions in the original MDP. This construction allows problem-specific operational structure, such as capacity constraints, integrality restrictions, and combinatorial coupling, to be handled by the decoder rather than by a customized policy architecture. Performance analysis suggests that the optimality gap can be decomposed into a structural approximation error and an algorithmic learning error. In the queueing network case study, the decoder yields a learned max-weight-type dispatching rule, and numerical results demonstrate that a standard PPO implementation outperform classical routing heuristics and ADP benchmarks considerably.

Several directions remain open for future research. First, this paper focuses on the infinite-horizon discounted-reward formulation. Many operational models are more naturally evaluated under a long-run average-cost criterion. Extending Bellman--Taylor score decoding to average-cost MDPs would require replacing the discounted continuation value with a relative value or bias function. Second, our framework assumes that the transition mechanism and a meaningful post-action representation are explicitly available. This is natural in many stylized OR models, but may be restrictive in data-driven settings where the dynamics are unknown. An important direction is to learn or adapt the post-action representation from data while retaining feasibility-preserving decoding. Third, the present implementation uses on-policy learning through PPO. Extending the framework to offline reinforcement learning would broaden its applicability and may improve sample efficiency, especially in systems where simulation is expensive or real-world exploration is limited.

\newpage 


\bibliographystyle{informs2014trsc} 
\bibliography{references} 


\newpage 
\end{document}